\documentclass[3p]{elsarticle}

\usepackage{amsfonts, amsmath, amssymb, amsthm}
\usepackage{enumerate}
\usepackage{tikz}
\usetikzlibrary{arrows}

\theoremstyle{definition}

\newcommand{\X}{\ensuremath{\mathbb{X}}}
\newcommand{\A}{\ensuremath{\mathbb{A}}}
\newcommand{\E}{\ensuremath{\mathbb{E}}}

\DeclareMathOperator*{\argmax}{arg\,max}

\begin{document}
\title{The Value Iteration Algorithm is Not Strongly Polynomial for
  Discounted Dynamic Programming}

\author{Eugene A. Feinberg}

\author{Jefferson Huang}


\address{Department of Applied Mathematics and Statistics, Stony Brook University, Stony Brook, NY 11794-3600, USA}
\begin{frontmatter}
  \begin{abstract}
    This note provides a simple example demonstrating that, if exact
    computations are allowed, the number of iterations required
    for the value iteration algorithm to find an optimal policy for
    discounted dynamic programming problems may grow arbitrarily
    quickly with the size of the problem. In particular, the number of
    iterations can be exponential in the number of actions. Thus, unlike policy iterations, the
    value iteration algorithm is not strongly polynomial for
    discounted dynamic programming.
  \end{abstract}
\end{frontmatter}

\section{Introduction}
\label{sec:introduction}

Value iterations, policy iterations, and linear programming are
three major methods for computing optimal policies for Markov
Decision Processes (MDPs) with expected total discounted rewards
\cite{kallenberg2002}, \cite[Chapter 6]{puterman1994}, also known
under the name of discounted dynamic programming. As is
well-known, policy iterations can be viewed as implementations of
the simplex method applied to one of the two major linear programs
used to solve MDPs; see e.g.\ \cite{kallenberg2002}, \cite[Section
6.9]{puterman1994}. Ye \cite{ye2011} proved that policy iterations
are strongly polynomial when the discount factor is fixed. This
note shows value iterations may not be strongly polynomial.

For value iteration, the best known upper bound on the required number
of iterations was obtained by Tseng \cite{tseng1990} (see also Littman
\cite{littman1995} and Ye \cite{ye2011}), and is a polynomial in the
number of states $n$, the number of actions $m$, the number of bits
$B$ needed to write down the problem data, and $(1 - \beta)^{-1}$,
where $\beta \in (0, 1)$ is the discount factor. Since the number of
arithmetic operations needed per iteration is at most a constant times
$n^2m$, this means that the value iteration algorithm is weakly
polynomial if the discount factor is fixed.

This note provides a simple example that demonstrates that, if
exact computations are allowed, the number of operations performed
by the value iteration algorithm can grow arbitrarily quickly as a
function of the total number of available actions. In particular,
the running time can be exponential with respect to the total
number of actions $m$. Thus, unlike policy iterations, value
iterations are not strongly polynomial.

\section{Example}
\label{sec:example}

Consider an arbitrary increasing sequence $\{M_i\}_{i = 1}^{\infty}$
of natural numbers. Let the state space be $\X = \{1, 2, 3\}$, and for
a natural number $k$ let the action space be $\A = \{0, 1, \dots,
k\}$. Let $\A(1) = \A$, $\A(2) = \{0\}$ and $\A(3) = \{0\}$ be the
sets of actions available at states 1, 2, and 3, respectively. The
transition probabilities are given by $p(2 | 1, i) = p(3 | 1, 0) = p(2
| 2, 0) = p(3 | 3, 0) = 1$ for $i = 1, \dots, k$. Finally, the
one-step rewards are given by $r(1, 0) = r(2, 0) = 0$, $r(3, 0) = 1$,
and
\begin{displaymath}
r(1, i) = \frac{\beta}{1 - \beta} (1 - \exp(-M_i)), \quad i = 1,
\dots, k.
\end{displaymath}
Figure~\ref{fig:k2} below illustrates such an MDP
for $k = 2$.
\begin{figure}[h]
  \centering
  \includegraphics[scale=1]{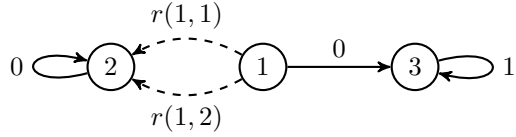}
  \caption{Diagram of the MDP for $k=2.$ The solid arcs correspond to transitions associated with
action 0, and dashed arcs correspond to the remaining actions. The
number next to each arc is the reward associated with the
corresponding action.}
\label{fig:k2}
\end{figure}

\subsection{Discounted-reward criterion}
\label{sec:disc-reward-crit}

Here we are interested in maximizing expected infinite-horizon
discounted rewards. In particular, a \textit{policy} is a mapping
$\phi: \X \rightarrow \A$ such that $\phi(x) \in \A(x)$ for each
$x \in \X$. It is possible to consider more general policies, but
for infinite-horizon discounted MDPs with finite state and action
sets it is sufficient to consider only policies of this form; see
e.g.\ \cite[p. 154]{puterman1994}. Let $F$ denote the set of all
policies. Also, given an initial state $x \in \X$, let
$\mathbb{P}^\phi_x$ denote the probability distribution on the set
of possible histories $x_0 a_0 x_1 a_1 \dots$ of the process under
the policy $\phi$ with $x_0 = x$, and let $\E^\phi_x$ be the
expectation operator associated with $\mathbb{P}^\phi_x$. Then the
expected total discounted reward earned when the policy $\phi$ is
used starting in state $x \in \X$ is
\begin{displaymath}
  v_\beta(x, \phi) = \E^\phi_x \sum_{t = 0}^{\infty} \beta^t r(x_t, a_t).
\end{displaymath}
The goal is to find an \textit{optimal policy}, i.e. a policy $\phi^*$
such that $v_\beta(x, \phi^*) = \sup_{\phi \in F} v_\beta(x,
\phi)$ for all $x \in \X$. It is well-known that if $\X$ and $\A$ are finite, then an
optimal policy exists; see e.g.\ \cite[p. 154]{puterman1994}.

For the above described MDP each policy is defined by an action
selected at state 1. Note that if action $i \in \{1, \dots, k\}$
is selected, then the total discounted reward starting from state
1 is  $r(1, i)$; if action 0 is selected, the corresponding total
discounted reward is $\beta/(1 - \beta)$. Since
\begin{displaymath}
  r(1, i) = \frac{\beta}{1 - \beta} (1 - \exp(-M_i)) < \frac{\beta}{1
    - \beta}
\end{displaymath}
for each $i = 1, \dots, k$, action 0 is the unique optimal action in
state 1.

\subsection{Running time of value iterations}
\label{sec:value-iteration}

We are interested in obtaining the optimal policy using \textit{value
  iteration}. In particular, set $V_0 \equiv 0$, and for each $x \in
\X$ and $j = 0, 1, \dots$ let
\begin{displaymath}
  V_{j + 1}(x) = \max_{a \in \A(x)} \{r(x, a) + \beta \sum_{y \in \X}
  p(y | x, a) V_j(y)\},
\end{displaymath}
and
\begin{displaymath}
  \phi^{j + 1}(x) \in \argmax_{a \in \A(x)} \{r(x, a) + \beta \sum_{y
    \in \X} p(y | x, a) V_j(y)\}.
\end{displaymath}
Since the numbers $M_i$ increase in $i$, for $j = 0, 1, \dots$
\begin{align*}
  V_{j + 1}(1) &= \max\left\{\frac{\beta(1 - \beta^j)}{
      1 - \beta}, \frac{\beta(1 - \exp(-M_k))}{1 - \beta} \right\}, \\
  V_{j + 1}(2) &= 0, \\
  V_{j + 1}(3) &= \frac{1 - \beta^{j + 1}}{1 - \beta},
\end{align*}
which means that
\begin{displaymath}
  \phi^{j + 1}(1) =
  \begin{cases}
    k, &\quad \text{if} \ j < M_k / (-\ln \beta), \\
    0, &\quad \text{if} \ j \geq M_k / (-\ln \beta).
  \end{cases}
\end{displaymath}
Hence more than $M_k / (-\ln \beta)$ iterations are needed to
select the optimal action 0 in state 1. In particular, since $m =
k + 3,$ for $M_k = e^{k + 3},$ $k=1,2,\ldots,$ more than $e^m /
(-\ln \beta)$ iterations are required to obtain the optimal
policy.


\newcommand{\noopsort}[1]{} \newcommand{\printfirst}[2]{#1}
  \newcommand{\singleletter}[1]{#1} \newcommand{\switchargs}[2]{#2#1}

\end{document}